\documentclass{article}

\usepackage{arxiv}

\usepackage[utf8]{inputenc} % allow utf-8 input
\usepackage[T1]{fontenc}    % use 8-bit T1 fonts
\usepackage{hyperref}       % hyperlinks
\usepackage{url}            % simple URL typesetting
\usepackage{booktabs}       % professional-quality tables
\usepackage{amsfonts}       % blackboard math symbols
\usepackage{nicefrac}       % compact symbols for 1/2, etc.
\usepackage{microtype}      % microtypography
\usepackage{lipsum}
\usepackage{graphicx}
\usepackage{cite}
\usepackage{amsmath,amssymb,amsfonts}
\usepackage{algorithmic}
\usepackage{graphicx}
\usepackage{textcomp}
\usepackage{xcolor}
\usepackage{multirow}
\usepackage{booktabs}
\usepackage{array}
\usepackage{graphicx} % For \resizebox
\usepackage{amsmath}
\usepackage{subcaption}
\usepackage{csquotes}
\usepackage{enumitem}
\usepackage{hyperref}
\usepackage{cleveref}
\usepackage{balance}
\usepackage[ruled,vlined,linesnumbered]{algorithm2e}
\usepackage{xurl}
\graphicspath{ {./images/} }

\begin{document}

\thispagestyle{empty}
{\Huge \textbf{IEEE Copyright Notice}}\\[1em]
{\large
\noindent
\textcopyright~2024 IEEE. Personal use of this material is permitted. Permission from IEEE must be obtained for all other uses, in any current or future media, including reprinting/republishing this material for advertising or promotional purposes, creating new collective works, for resale or redistribution to servers or lists, or reuse of any copyrighted component of this work in other works.

DOI: \href{https://doi.org/10.1109/ICCIT64611.2024.11022335}{10.1109/ICCIT64611.2024.11022335}
}
\newpage

\fancypagestyle{titlepage}{
  \fancyhf{}
  \fancyhead[C]{\footnotesize This work has been accepted for publication in 2024 27th International Conference on Computer and Information Technology (ICCIT).\\
  The final published version is available via IEEE Xplore.\\
  DOI: \href{https://doi.org/10.1109/ICCIT64611.2024.11022335}{10.1109/ICCIT64611.2024.11022335}
  }
  \renewcommand{\headrulewidth}{0pt}
}

\title{An Approach Towards Identifying Bangladeshi Leaf Diseases through Transfer Learning and XAI}

\author{
 Faika Fairuj Preotee \\
  Department of Computer Science and Engineering\\
  Ahsanullah University of Science and Technology\\
  Dhaka,Bangladesh\\
  \texttt{faikafairuj2001@gmail.com} \\
  %% examples of more authors
    \And
 Shuvashis Sarker \\
  Department of Computer Science and Engineering\\
  Ahsanullah University of Science and Technology\\
  Dhaka,Bangladesh\\
  \texttt{shuvashisofficial@gmail.com} \\
   \And
Shamim Rahim Refat \\
  Department of Computer Science and Engineering\\
  Ahsanullah University of Science and Technology\\
  Dhaka,Bangladesh\\
  \texttt{n.a.refat2000@gmail.com} \\
  \And
Tashreef Muhammad \\
  Department of Computer Science and Engineering\\
  Southeast University\\
  Dhaka,Bangladesh\\
  \texttt{tashreef.muhammad@gmail.com}\\
\And
 Shifat Islam \\
  Department of Computer Science and Engineering\\
  Bangladesh University of Engineering and Technology\\
  Dhaka,Bangladesh\\
  \texttt{shifat.islam.buet@gmail.com} \\
}

\maketitle
\thispagestyle{titlepage}

\begin{abstract}
Leaf diseases are harmful conditions that affect the health, appearance and productivity of plants, leading to significant plant loss and negatively impacting farmers' livelihoods. These diseases cause visible symptoms such as lesions, color changes, and texture variations, making it difficult for farmers to manage plant health, especially in large or remote farms where expert knowledge is limited. The main motivation of this study is to provide an efficient and accessible solution for identifying plant leaf diseases in Bangladesh, where agriculture plays a critical role in food security. The objective of our research is to classify 21 distinct leaf diseases across six plants using deep learning models, improving disease detection accuracy while reducing the need for expert involvement. Deep Learning (DL) techniques, including CNN and Transfer Learning (TL) models like VGG16, VGG19, MobileNetV2, InceptionV3, ResNet50V2 and Xception are used. VGG19 and Xception achieve the highest accuracies, with 98.90\% and 98.66\% respectively. Additionally, Explainable AI (XAI) techniques such as GradCAM, GradCAM++, LayerCAM, ScoreCAM and FasterScoreCAM are used to enhance transparency by highlighting the regions of the models focused on during disease classification. This transparency ensures that farmers can understand the model's predictions and take necessary action. This approach not only improves disease management but also supports farmers in making informed decisions, leading to better plant protection and increased agricultural productivity.
\end{abstract}


\section{Introduction}
Plant leaf disease affects the health and appearance of leaves, significantly impacting plant production. % These diseases are identified by visible changes such as color, texture, and the appearance of lesions. 
%Molecular, serological, and microbiological methodologies offer accurate but labour-intensive answers.
Conventional visual inspection necessitates specialised expertise, rendering it labour-intensive and expensive, particularly on extensive farms or in isolated regions. The global population is expected to attain 9.1 Billion by 2050 \cite{bruinsma2009resource}. So, efficient disease detection and management practices are critical for ensuring food security.

Bangladesh, as one of the most densely populated countries, relies heavily on agriculture for economic stability. Managing plant health is challenging due to limited information on plant illness. With limited land and a large population, efficient disease management is crucial for food security. Automated plant disease detection systems will further support agricultural sustainability.

Transfer Learning (TL) techniques have been extensively utilized in plant disease detection, owing to their capacity to utilize pre-trained models for accurately identifying and classifying plant diseases. TL automates disease detection, minimizing the necessity for expert involvement and delivering quick information for effective plant management. Fine-tuned pre-trained models are capable of classifying diverse plant diseases and frequently surpass state-of-the-art models in performance.

% In previous studies, the focus has been primarily on detecting plant diseases in a single variant of plants and no extensive research has been conducted on datasets covering multiple plant types. In this study, the gap is addressed by analyzing various leaf diseases affecting different types of plants. The challenges mentioned are tackled in this paper through the following key contributions.
In former times, much of the attention has been given to plant disease detection in specific plant types. However, this study expands the scope by focusing on multiple plant species, analyzing a diverse range of leaf diseases. Through this comprehensive approach, the study addresses key challenges related to disease classification and model interpretability. These challenges are tackled through the following key contributions:
\begin{enumerate}[label=\roman*.]
    \item To implement a deep learning-based CNN method to classify various leaf diseases affecting multiple plant species, addressing key challenges in plant pathology.
    \item To apply fine-tuned transfer learning models with pre-trained weights for precise and efficient identification of leaf diseases.
    \item To utilize explainable AI (XAI) techniques to enhance model interpretability by highlighting key regions for disease detection, comparing multiple XAI methods.
    \item To identify the optimal classification approach, aiming for high accuracy and suitability for practical agricultural use.
\end{enumerate}

The rest of the paper is organized as follows: Section \ref{Related Work} discusses the related works. The dataset used for the study is detailed in Section \ref{Dataset}. Section \ref{Methodology} outlines the proposed methodology, while the experimental setup and result analysis are presented in Section \ref{Experimental setup Result Analysis}. Explainable AI techniques are explored in Section \ref{Explainable AI}. Finally, Section \ref{Conclusion Future Work} concludes the paper with a discussion of the conclusions and future work.

\section{Related Work}
\label{Related Work}

\subsection{Deep Learning (DL) Model}
DL models evaluate huge datasets to precisely identify plant diseases, enhancing both speed and reliability of the process. According to this, Hossain et al. \cite{hossain2021plant} presented three depth-wise separable convolutional models (S-modified, S-reduced, and S-extended MobileNet) for the identification of plant leaf diseases utilizing Adaptive Centroid-based Segmentation (ACS) on datasets such as PlantVillage (PV), Kaggle, IRRI and BRRI. The S-modified MobileNet attained an accuracy of 99.55\% and an F1-score of 97.07\%. For exploring further more, Vallabhajosyula et al. \cite{vallabhajosyula2022transfer} created a Deep Ensemble Neural Network (DENN) for diagnosing plant diseases with the PV dataset, achieving an accuracy of 99.99\%, surpassing models such as ResNet and DenseNet. The publicly available PV \cite{hughes2015open} dataset had been used in multiple research works. Shewale et al. \cite{shewale2023high} developed a CNN for the early identification of plant diseases, attaining an accuracy of 99.81\% through preprocessing, segmentation and data augmentation using images from Jalgaon fields and the PV dataset. Ahmad et al.\cite{ahmad2021plant} employed MobileNet V3 Large with incremental transfer learning for plant disease identification, attaining 99.69\% accuracy on the PV dataset and 99\% on the Pepper dataset via data augmentation and class balancing. Akther et al. \cite{akther2024comparative} employed CNN, EfficientNet and ResNet to detect illnesses in grape and potato leaves from Sonaimuri, Bangladesh, with EfficientNet attaining 97\% accuracy through TL and preprocessing techniques such as greyscale conversion and noise reduction.

\subsection{Explainable AI (XAI)}
\par To enhance transparency and interpretability, Explainable AI (XAI) has gained significance, with model visualization being a crucial technique for explaining DL. Following this approach,  Bhandari et al.\cite{bhandari2023botanicx} created a tomato leaf disease detection model with EfficientNetB5, employing photos sourced from the Kaggle dataset. The model attained an accuracy of 99.07\% and employed GradCAM and LIME for interpretability, outperforming models such as MobileNet and VGG16. Moreover, Mahmud et al.\cite{mahmud2023explainable} employed EfficientNetB3, Xception and MobileNetV2 for the diagnosis of tomato leaf diseases, achieving an accuracy of 99.30\% with EfficientNetB3, while utilizing GradCAM and Saliency Maps for interpretability. Further enhancing interpretability, Nahiduzzaman et al.\cite{nahiduzzaman2023explainable} introduced a PDS-CNN for the classification of Bangladeshi mulberry leaf diseases, attaining an accuracy of 95.05\% for three classes and 96.06\% for binary classification. The research highlighted XAI with SHAP to explain the model's decisions. Likewise, Wei et al.\cite{wei2022explainable} employed DL models, specifically ResNet-CBAM, for the classification of fruit leaf diseases utilising data from the PV, Google and AI Challenger datasets. The model attained an accuracy of 99.89\%, using XAI methodologies such as SmoothGrad, LIME and GradCAM, with GradCAM. Building on this, Batchuluun et al.\cite{batchuluun2022deep} introduced the PlantDXAI methodology employing CNN-16 for the identification of plant diseases in rose and rice crops, attaining an accuracy of 98.55\% for roses and 90.04\% for paddy crops, utilizing CAM to emphasise damaged regions during classification.

\begin{table}[ht]
\centering
\caption{Summary  of the reviewd work}
\label{Table-ComparisonXAI}
\renewcommand{\arraystretch}{1.1}
\begin{tabular}{|c|c|c|c|}
\hline
\textbf{Ref. No.} & \textbf{Proposed Method} & \textbf{Accuracy} & \textbf{Use of XAI} \\ \hline
\cite{hossain2021plant} & \begin{tabular}[c]{@{}c@{}}MobileNet \\ (Modified)\end{tabular} & 99.55\% & NO \\ \hline
\cite{vallabhajosyula2022transfer} & \begin{tabular}[c]{@{}c@{}}Deep Evolving \\ Neural Net\end{tabular} & 99.99\% & NO \\ \hline
\cite{shewale2023high} & \begin{tabular}[c]{@{}c@{}}Convolutional \\ Neural Network\end{tabular} & 99.81\% & NO \\ \hline
\cite{ahmad2021plant}& MobileNetV3 & 99.69\% & NO \\ \hline
\cite{akther2024comparative} & EfficientNet & 97.00\% & NO \\ \hline
\cite{bhandari2023botanicx} & \begin{tabular}[c]{@{}c@{}}EfficientNet \\ B5\end{tabular} & 99.07\% & \begin{tabular}[c]{@{}c@{}}YES\\ (Grad-CAM, \\ LIME)\end{tabular} \\ \hline
\cite{mahmud2023explainable} & \begin{tabular}[c]{@{}c@{}}EfficientNet \\ B3\end{tabular} & 99.30\% & \begin{tabular}[c]{@{}c@{}}YES\\ (Grad-CAM)\end{tabular} \\ \hline
\cite{nahiduzzaman2023explainable} & \begin{tabular}[c]{@{}c@{}}PDS-Convolutional \\ Network\end{tabular} & 96.06\% & \begin{tabular}[c]{@{}c@{}}YES\\ (SHAP)\end{tabular} \\ \hline
\cite{wei2022explainable} & \begin{tabular}[c]{@{}c@{}}ResNet with \\ CBAM\end{tabular} & 99.89\% & \begin{tabular}[c]{@{}c@{}}YES\\ (Grad-CAM)\end{tabular} \\ \hline
\cite{batchuluun2022deep} & \begin{tabular}[c]{@{}c@{}}CNN-16 \end{tabular} & \begin{tabular}[c]{@{}c@{}}98.55\% \\ (Roses)\\ 90.04\% \\ (Paddy Crops)\end{tabular} & \begin{tabular}[c]{@{}c@{}}YES\\ (CAM)\end{tabular} \\ \hline
\end{tabular}
\end{table}
\par Table \ref{Table-ComparisonXAI} provides the  summary of our  reviewd work. Here, Ref. No. denotes reference number of the related works. After reviewing existing research on plant leaf datasets, it is found that no significant work has been conducted on the pure Bangladeshi dataset %\textit{Plant Leaf Freshness and Disease Detection Dataset From Bangladesh}, 
particularly in the classification of multiple diseases across various plant types. This study aims to address this gap by utilizing advanced TL models in combination with XAI techniques to classify plant diseases and elucidate the model's decision-making process.
\section{Dataset}
\label{Dataset}
The dataset which is used in this study is a publicly available dataset from \textbf{\textit{Mendeley Data}} \cite{plant_leaf_dataset} proposed by Mahamudul Hasan et al. \cite{hasan2024comprehensive}. It consists of 12,786 high-quality images of plant leaves, carefully organized to support advanced ML and DL models for plant disease detection. 

\begin{figure}[h!]
    \centering
    \includegraphics[width=\linewidth]{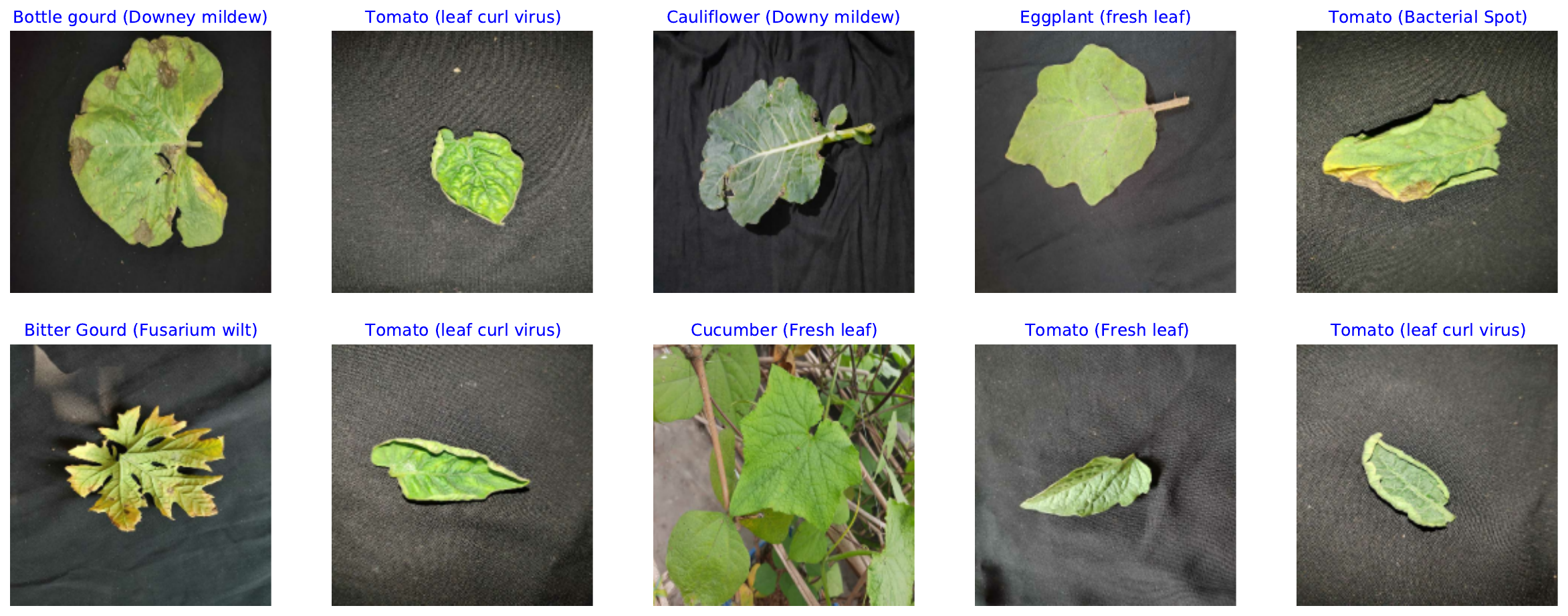}
    \caption{Sample Images of the Dataset}
    \label{Figure 1}
\end{figure}

\Cref{Figure 1} shows some sample images of the dataset. The dataset includes images from six different plant types: \textit{Bitter Gourd, Bottle Gourd, Cauliflower, Eggplant, Cucumber and Tomato}, each categorized into healthy and various disease-affected classes. Diseases identified include Downy mildew, Mosaic virus, Fusarium wilt, Black rot, Bacterial spot and Tomato leaf curl virus among others. 

\begin{figure}[h!]
    \centering
    \includegraphics[width=\linewidth]{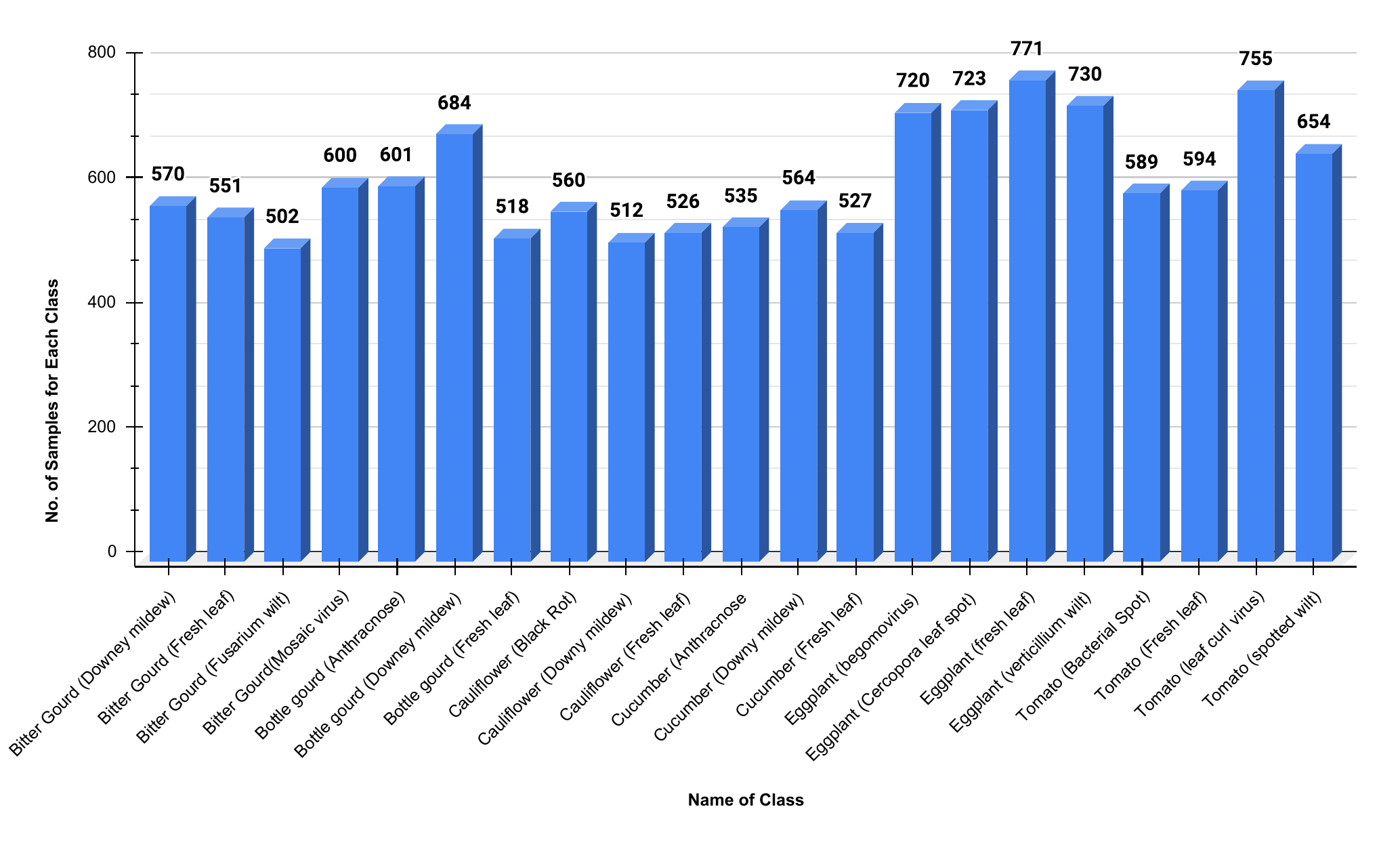}
    \caption{Different Class Distribution of the Dataset}
    \label{Figure 2}
\end{figure}

Data collection is carried out in different agricultural fields in{\textit{Kathalkandi, Nasirnagar and Brahmanbaria, Bangladesh}}. The collection process involves close collaboration with local farmers to accurately identify and classify the plants and diseases. The leaves are carefully categorized, ensuring no mismatches between plant types and the images are captured using high-quality techniques, with a black background to reduce noise and enhance the focus on the leaf details. \Cref{Figure 2} shows a picture of how the different classes in the sample have been distributed out.\

\section{Methodology}
\label{Methodology}
\subsection{Data Preprocessing}
\begin{figure}[h!]
    \centering
    \includegraphics[width=\linewidth]{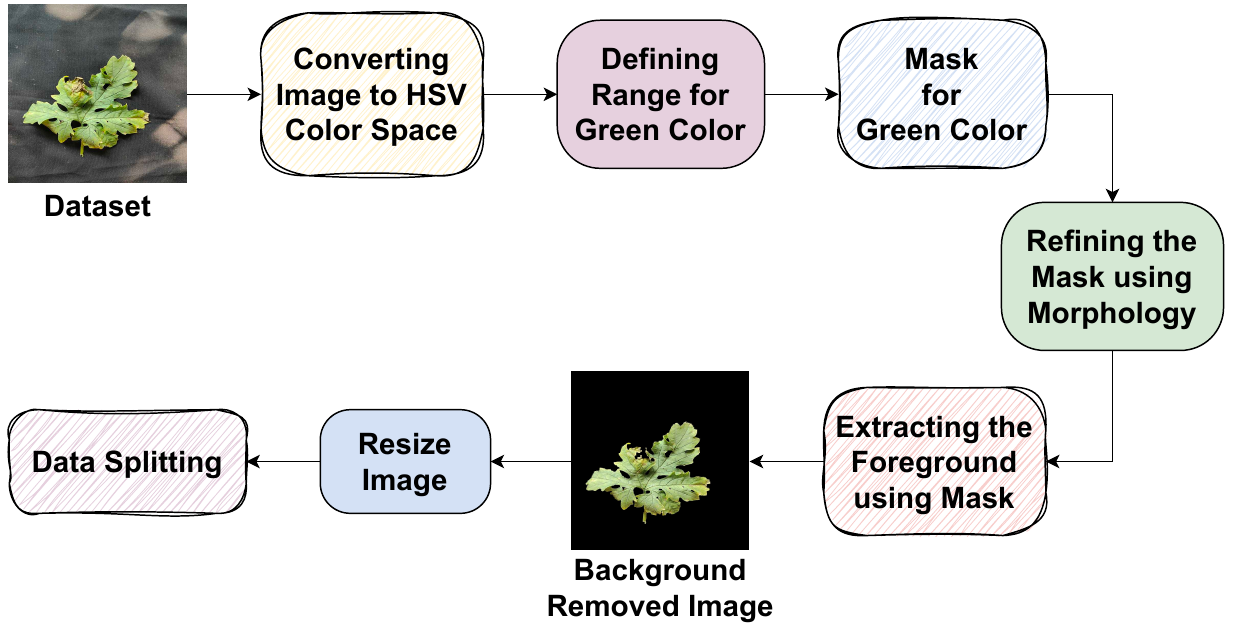}
    \caption{Data Preprocessing Technique}
    \label{Figure 3}
\end{figure}

The initial dataset consists of six classes, each containing multiple disease categories along with normal leaf images for the respective plant types. These six categories are generalized into 21 classes representing various diseases and healthy leaves. \Cref{Figure 3} shows the overall data preprocessing steps, including background removal and image resizing.

In the data preprocessing stage, background removal is utilized to separate the leaf from it's surrounding background. The procedure is started with the transformation of the image into the HSV color space to improve color-based segmentation. The green color range for the leaves is defined, with the lower limit set at 25, 40, 40 and the upper limit set at 90, 255, 255. This facilitates accurate identification of the leaf's green regions. A mask is subsequently generated to separate the leaf from the background. Morphological operation, including opening and closing, are utilized to enhance the mask and eliminate noise or minor artifacts. The final phase consists of eliminating the leaf (foreground) utilizing the mask. The original dataset photos, measuring 700x700 pixels, have been reduced to 128x128 pixels to standardize the input dimensions for the DL models, thereby decreasing computational expenses while preserving the necessary image quality for precise disease diagnosis. The dataset is partitioned into training, validation and test sets in an 80:10:10 ratio, designating 80\% for training, 10\% for validation and 10\% for testing. This equitable division facilitates the correct training and evaluation of the models. 

\subsection{Model Architecture}
\begin{figure}[h!]
    \centering
    \includegraphics[width=\linewidth]{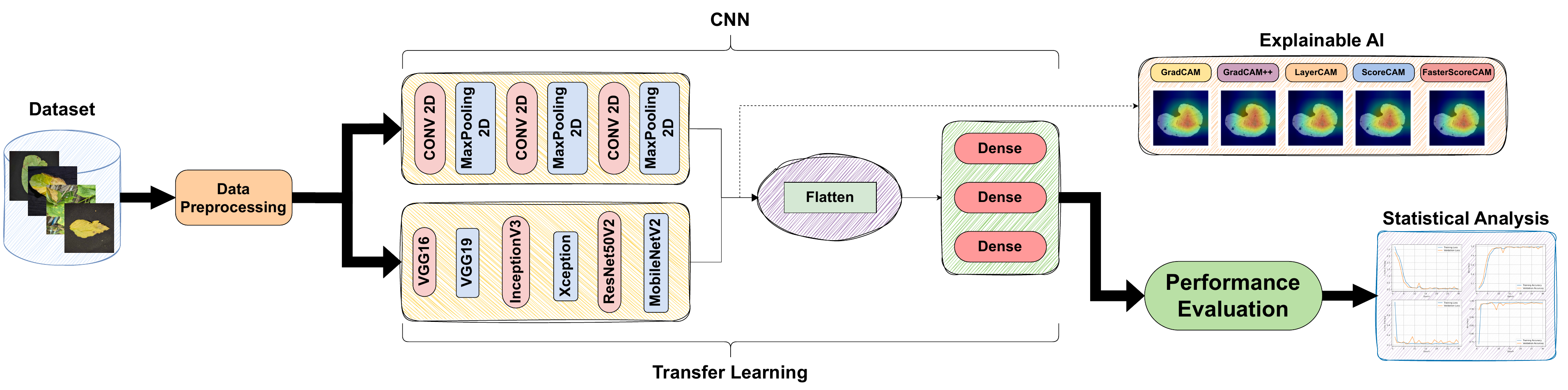}
    \caption{Workflow of Proposed Method}
    \label{Figure 4}
\end{figure}

\Cref{Figure 4} shows the workflow of the proposed method. This study employs an architecture for plant disease classification that employs CNN, renowned for their capacity to identify complex patterns in images via convolutional, pooling and fully connected layers. Convolutional layers collect essential information from input images, whereas activation functions such as ReLU introduce non-linearity to improve model effectiveness. Multiple pre-trained models based on CNN architectures, such as VGG16, VGG19, ResNet50V2, InceptionV3, Xception and MobileNetV2 are employed to enhance classification accuracy. These models evaluate images at different scales: VGG employs 3x3 filters for intricate feature extraction, ResNet50 employs residual connections to reduce vanishing gradients, InceptionV3 captures multi-scale features, Xception enhances efficiency via depthwise separable convolutions and MobileNet optimises performance for mobile and resource limited devices. The pre-trained models have established a strong basis for the effective classification of plant diseases\cite{salehi2023study}.

This study incorporates a range of models to effectively address various challenges in plant disease classification. The CNN initiates with three Conv2D layers, each followed by a MaxPooling2D layer to extract and reduce feature dimensionality. The Conv2D layers begin with 32 filters, increase to 128 and subsequently decrease to 64, so enhancing the model's capacity to capture diverse image properties. Subsequent to convolutional processing, the data is flattened and transmitted through two fully connected layers comprising 1024 and 512 neurons respectively, each followed by dropout layers to mitigate overfitting. A concluding dense layer, employing a softmax activation function, categorized the data into 21 classes. Furthermore, many pre-trained models, including VGG16, VGG19, InceptionV3, Xception, ResNet50V2 and MobileNetV2 are employed with their weights initialized from ImageNet. The pre-trained models are fine-tuned using the plant disease dataset to adapt their learnt features for this specific task. The model architecture comprises a flatten layer, dense layers containing 1024 and 512 neurons, dropout layers and a softmax output layer for classification purposes. Fine-tuning the pre-trained models using the training dataset enhances the performance and decreases the training duration by utilizing their inherent capabilities based on ImageNet weights.
\section{Experimental setup \& Result Analysis}
\label{Experimental setup Result Analysis}

\subsection{Experimental setup}
The methods are implemented on a computer system equipped with an Intel Core i5 13400F processor, an NVIDIA GeForce RTX 3060 12GB GPU and 16GB of DDR4 RAM. The models are executed using Keras \cite{chollet2015keras}, a high-level neural networks API, with TensorFlow \cite{tensorflow2015-whitepaper} as the backend. Each model runs five times to ensure robustness, with the optimal results recorded for analysis. All models are trained with a \textbf{Learning Rate(LR)} of \textbf{0.0001} and executed for \textbf{30 epochs}, except the \textbf{CNN}, which is trained for \textbf{60 epochs}. \textbf{SparseCategoricalCrossentropy} serve as the loss function, while the \textbf{Adam} optimizer is utilized to attain optimal performance across all models.

\subsection{Result Analysis}
Table \ref{tab:Table1} displays the performance data for several DL models, including \textbf{CNN, Xception, InceptionV3, ResNet50V2, VGG16, VGG19} and \textbf{MobileNetV2}. The models are evaluated according to the overall score.

\begin{table}[htbp]
\centering
\caption{Performance Metrics for Different Models}
\label{tab:Table1}
\renewcommand{\arraystretch}{1.2}
\setlength\tabcolsep{6pt} % Adjust column spacing
\small % Reduce font size to fit within \textwidth
\begin{tabular}{|c|c|c|c|c|c|c|}
\hline
\textbf{Model} & \textbf{Epoch} & \textbf{LR} & \textbf{Accuracy} & \textbf{Precision} & \textbf{Recall} & \textbf{F1-score} \\ \hline
CNN            & 60             & \multirow{7}{*}{0.0001} & 0.92175         & 0.92             & 0.92           & 0.92             \\ \cline{1-2} \cline{4-7}
\textbf{Xception} & \multirow{6}{*}{30} &                     & \textbf{0.98669} & \textbf{0.99}    & \textbf{0.98}  & \textbf{0.98}    \\ \cline{1-1} \cline{4-7}
InceptionV3    &                &                          & 0.97887         & 0.98             & 0.98           & 0.98             \\ \cline{1-1} \cline{4-7}
ResNet50V2     &                &                          & 0.97417         & 0.97             & 0.97           & 0.97             \\ \cline{1-1} \cline{4-7}
VGG16          &                &                          & 0.98356         & 0.98             & 0.98           & 0.98             \\ \cline{1-1} \cline{4-7}
\textbf{VGG19} &                &                          & \textbf{0.98904} & \textbf{0.99}    & \textbf{0.99}  & \textbf{0.99}    \\ \cline{1-1} \cline{4-7}
MobileNetV2    &                &                          & 0.97809         & 0.98             & 0.98           & 0.98             \\ \hline
\end{tabular}
\end{table}

\textbf{Xception} and \textbf{VGG19} display superior performance, attaining the greatest accuracies of \textbf{98.66\%} and \textbf{98.90\%} respectively. \textbf{VGG19} model attain the best precision, recall and F1-score which is 0.99 demonstrating their exceptional efficacy in accurately recognizing and classifying plant diseases. Whereas, Xception has also achieved a precision of 0.99, with a recall and F1-score of 0.98.

\begin{figure}[h]
    \centering
    \begin{subfigure}[b]{0.46\textwidth}
        \centering
        \includegraphics[width=\linewidth]{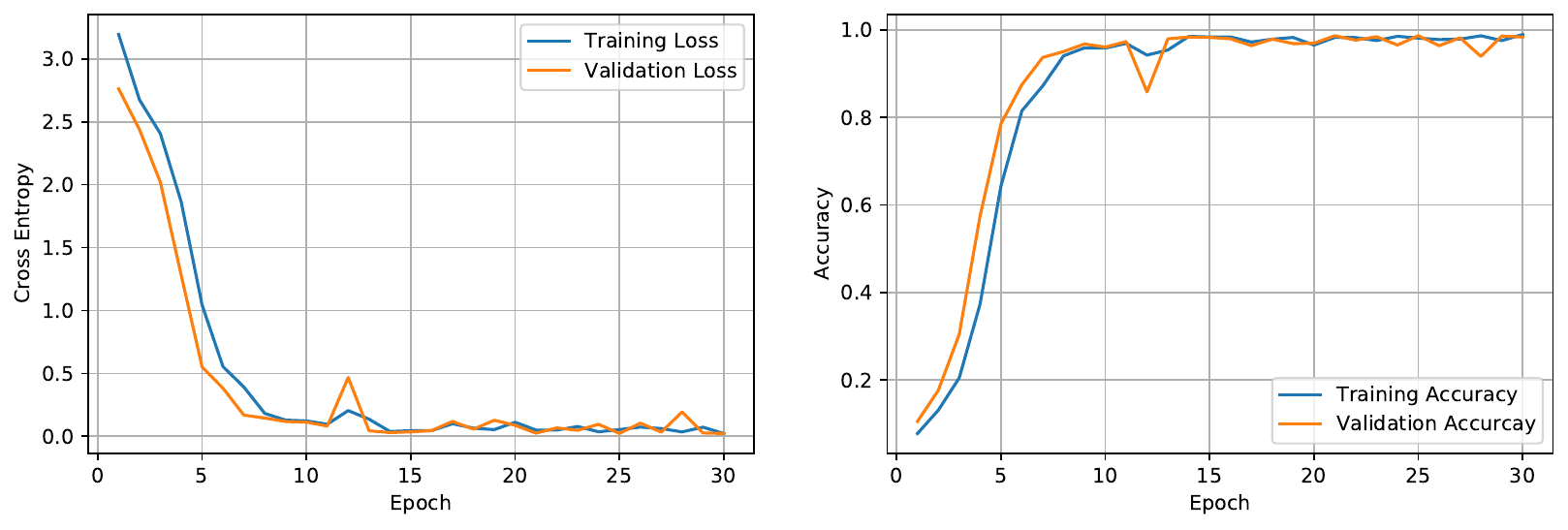} % Replace with your image file
        \caption{VGG19}
        \label{fig:subfig1}
    \end{subfigure}
    \hspace{0.02\textwidth} % Adjust the spacing between the subfigures if needed
    \begin{subfigure}[b]{0.46\textwidth}
        \centering
        \includegraphics[width=\linewidth]{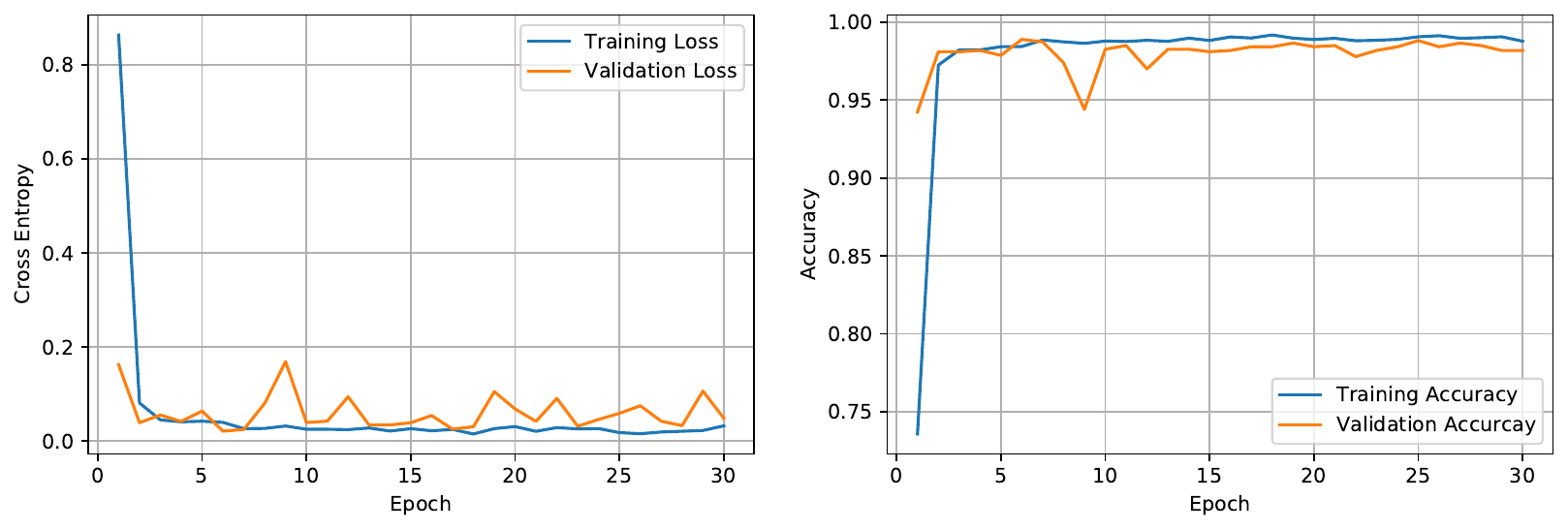} % Replace with your image file
        \caption{Xception}
        \label{fig:subfig2}
    \end{subfigure}
    %\caption{Comparison of training and validation loss and accuracy for two models.}
    \label{fig:comparison}
    \begin{subfigure}[b]{0.46\textwidth}
        \centering
        \includegraphics[width=\linewidth]{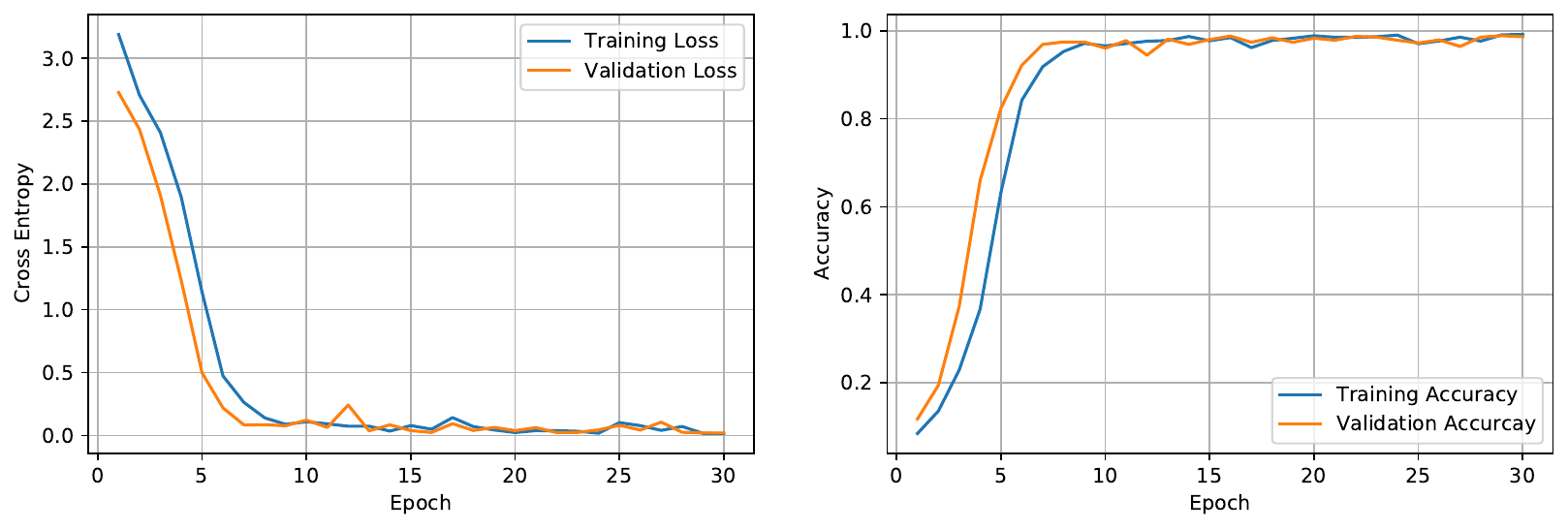} % Replace with your image file
        \caption{VGG16}
        \label{fig:subfig3}
    \end{subfigure}
    \hspace{0.02\textwidth} % Adjust the spacing between the subfigures if needed
    \begin{subfigure}[b]{0.46\textwidth}
        \centering
        \includegraphics[width=\linewidth]{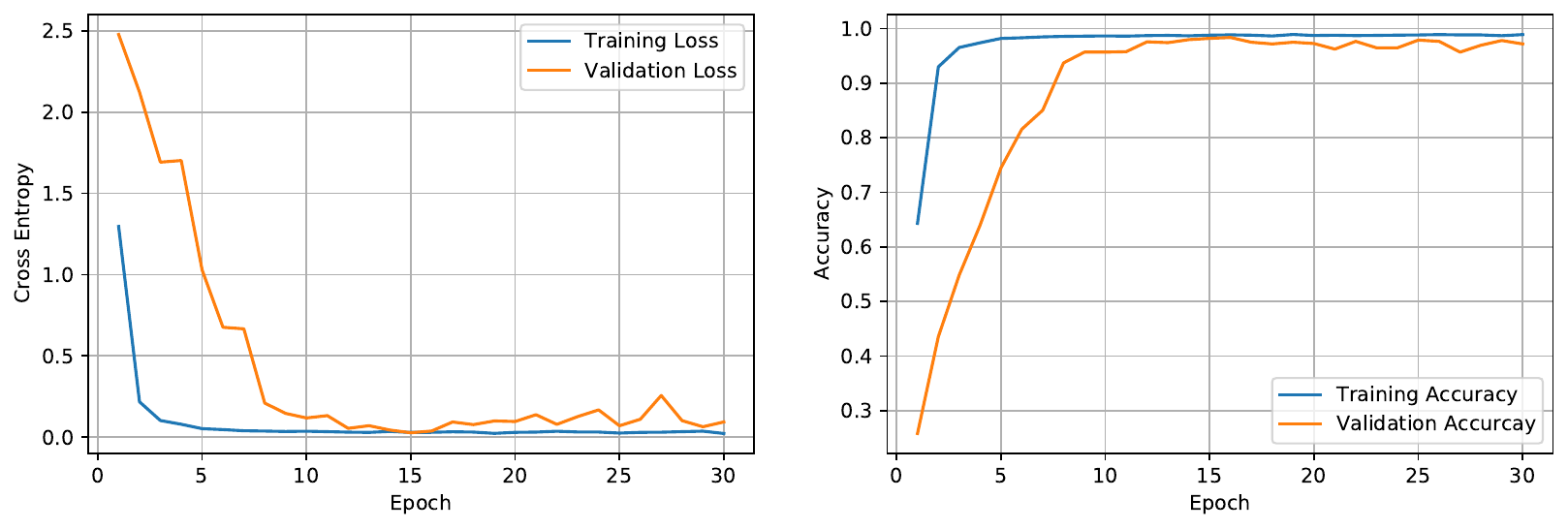} % Replace with your image file
        \caption{MobileNetV2}
        \label{fig:subfig4}
    \end{subfigure}
    \caption{Accuracy and Loss Curves for training and validation data in different models}
    \label{Figure 5}
\end{figure}

InceptionV3 and MobileNetV2 show exceptional performance, achieving accuracies 97.88\% and 97.80\% respectively along with consistently high precision, recall, f1- scores (0.98), indicating their robustness across several criteria. Resnet50V2 also perform  well, achieving an accuracy of 98.41\%, with a precision, recall and an F1-score of 0.97. VGG16 achieve an accuracy of 98.35\%, displaying a robust stability among precision, recall and F1-score at 0.98.

Although the CNN model attain a commendable accuracy of 92.17\%, along with precision, recall and F1-score of 0.92, it has fallen short in comparison to the more sophisticated models in the table, underscoring the improvements which are provided by deeper architectures and transfer learning methodologies.

\Cref{Figure 5} shows accuracy and loss curves for training and validation data for \textbf{VGG19, Xception, VGG16} and \textbf{MobileNetV2} demonstrate quick convergence, with all models attaining higher accuracy at an early stage. VGG19 and Xception exhibit the most consistent performance, with negligible divergence between training and validation curves, indicating effective generalization. VGG16 and MobileNetV2 exhibit commendable performance. However, they demonstrate a little slower stability in validation accuracy. All models demonstrate proficient learning with minimal overfitting.

\section{Explainable AI}
\label{Explainable AI}
XAI embraces methodologies that enhance the transparency of AI models by explaining their decision-making processes, especially within DL and TL frameworks. \textbf{GradCAM} is a pivotal XAI technique that offers visual explanations for CNNs by utilizing the gradients of the target class to identify the most significant regions in an image for prediction, thereby producing a heatmap that highlights these areas. It improves transparency and interpretability without necessitating modifications to the architecture, rendering it useful for applications like as image classification, captioning, and visual question answering (VQA) \cite{rafi2022deep}.

\begin{figure}[h!]
    \centering
    \includegraphics[width=\linewidth]{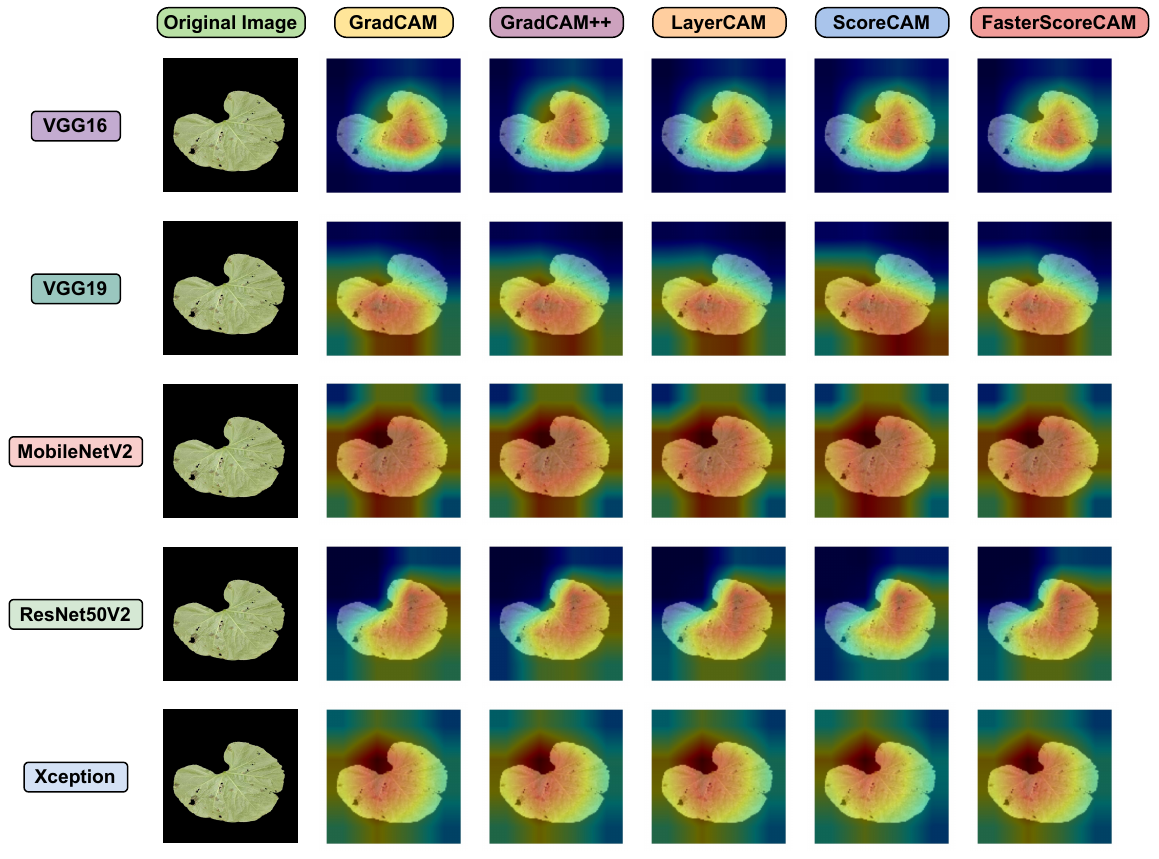}
    \caption{Comparative Analysis of Explainable AI Methods}
    \label{Figure 6}
\end{figure}

\textbf{GradCAM++}, an enhanced variation, resolves issues in localizing multiple instances of identical classes and more effectively defines objects by integrating second-order derivatives, hence improving heatmap accuracy. \textbf{ScoreCAM} presents an alternative methodology by evaluating the impact of activation maps by output score variations as the image changes, generating class-specific maps independent of gradients. This technique improves visual clarity, rendering it efficient for image classification by identifying crucial regions utilized by the model. \textbf{FasterScore-CAM} enhances this procedure by focussing on the most active channels within feature maps, hence decreasing computing expenses while preserving interpretability, which is especially advantageous for large datasets or complex models. Finally, \textbf{LayerCAM} improves interpretability by producing class activation maps from various layers of the CNN, rather than solely from the final convolutional layer, thereby offering more extensive insights into spatial localization and object attributes, which enhances object localization and classification precision.\cite{faria2024explainable}

\Cref{Figure 6} shows for various TL models representing consistent activation patterns across the leaf when analyzed using the five XAI approaches. To achieve this, the methods utilize feature maps particularly from the last or intermediate layers before the flattening layer.

In all applied TL models, GradCAM highlights large activations in critical regions of the leaf, but GradCAM++ produces more focused and accurate heatmaps. LayerCAM exhibits wider activations while concentrating on significant areas, whereas ScoreCAM distributes attention more evenly across the leaf surface. FasterScoreCAM, in contrast, concentrates on particular areas exhibiting greater confidence. Across the models, GradCAM and GradCAM++ emphasise key and pivotal regions, whereas LayerCAM maintains focus with broader coverage. Both ScoreCAM and FasterScoreCAM regularly exhibit focused attention on critical areas, indicating that the models typically recognise essentially identical regions of interest across various XAI methodologies. This work illustrates the significance of using the most appropriate XAI strategies customised for each transfer learning model to attain the most precise and thorough deep learning predictions for plant disease classification.

\section{Conclusion \& Future Work}
\label{Conclusion Future Work}
This study is conducted in the context of Bangladesh, seeks to support poor farmers by establishing an accurate and transparent system for diagnosing plant leaf illnesses with DL techniques. Through the implementation of TL, the model attains improved accuracy, enhancing disease detection efficiency and accessibility for resource constrained farmers. The implementation of XAI enhances the system's transparency, enabling farmers to readily discern the impacted regions on the plants and comprehend the rationale behind the model's forecasts. This can result in expedited and more efficient disease management, assisting farmers in safeguarding their crops and livelihoods, therefore enhancing total agricultural productivity in rural regions. 

Future work will focus on expanding the system to a broader scale, integrating it into practical applications for real-life use by farmers. Efforts will include working with more diverse datasets to improve the model's generalizability across various plant species and diseases. Additionally, implementing the system in real-world agricultural settings, potentially using IoT-based sensors for continuous monitoring, is a key goal. Further optimizations will focus on improving real-time performance on edge devices and enhancing model interpretability through expanded Explainable AI (XAI) techniques, ensuring ease of use for non-expert users.

\bibliographystyle{unsrt}  


\bibliography{references.bib}

\end{document}